\documentclass[twocolumn]{svjour3}          
\smartqed  
\usepackage{graphicx}
\usepackage{mathptmx}      
\usepackage[numbers]{natbib}

\usepackage{amsmath}
\usepackage{amssymb}
\usepackage{multirow}
\usepackage{setspace}
\usepackage{url}

\usepackage[normalem]{ulem}
\usepackage{color}

\newcommand{\blue}[1]{{\color{black} #1}}

\newcommand{\red}[1]{{\color{black} #1}}
\newcommand{\RR}{{\mathbb{R}}}

\begin{document}

\title{Revisiting convolutional neural network on graphs with polynomial approximations of Laplace-Beltrami spectral filtering}

\author{Shih-Gu Huang  \and  Moo K. Chung  \and   Anqi Qiu  \and  Alzheimer's Disease Neuroimaging Initiative}

\institute{
		Correspondence to Anqi Qiu (bieqa@nus.edu.sg)  \\
             		 \\        
		S.-G. Huang  \at
              Department of Biomedical Engineering, National University of Singapore, Singapore 117583  
           \and
          M.K. Chung \at
              Department of Biostatistics and Medical Informatics, University of Wisconsin, Madison, WI 53706, USA  
           \and            
          A. Qiu \at
              Department of Biomedical Engineering and the N.1 Institute for Health, National University of Singapore, Singapore, 117583  \\
}

\date{Received: date / Accepted: date}

\maketitle

\begin{abstract}
This paper revisits spectral graph convolutional neural networks (graph-CNNs) given in Defferrard (2016) and develops the Laplace-Beltrami CNN (LB-CNN) by replacing the graph Laplacian with the LB operator. We then define spectral filters via the LB operator on a graph. We explore the feasibility of Chebyshev, Laguerre, and Hermite polynomials to approximate LB-based spectral filters and define an update of the LB operator for pooling in the LB-CNN. We employ the brain image data from Alzheimer's Disease Neuroimaging Initiative (ADNI) and demonstrate the use of the proposed LB-CNN. Based on the cortical thickness of the ADNI dataset, we showed that the LB-CNN didn't improve classification accuracy compared to the spectral graph-CNN. The three polynomials had a similar computational cost and showed comparable classification accuracy in the LB-CNN or spectral graph-CNN. Our findings suggest that even though the shapes of the three polynomials are different, deep learning architecture allows us to learn spectral filters such that the classification performance is not dependent on the type of the polynomials or the operators (graph Laplacian and LB operator). 

\end{abstract}

\section*{Declarations}
\subsection*{Funding} 
\vspace*{-10pt} This research/project is supported by the National Science Foundation MDS-2010778, National Institute of Health R01 EB022856, EB02875, and National Research Foundation, Singapore under its AI Singapore Programme (AISG Award No: AISG-GC-2019-002). Additional funding is provided by the Singapore Ministry of Education (Academic research fund Tier 1; NUHSRO/2017/052/T1-SRP-Partnership/01), NUS Institute of Data Science. This research was also supported by the A*STAR Computational Resource Centre through the use of its high-performance computing facilities.

\vspace*{-10pt}
\subsection*{Conflicts of interest/Competing interests} 
\vspace*{-10pt} The authors declare that they have no conflict of interest.

\subsection*{Availability of data and material} 
\vspace*{-10pt} Data used in preparation of this article were obtained from the Alzheimer's Disease Neuroimaging Initiative (ADNI) database (\protect\url{adni.loni.ucla.edu}).

\vspace*{-10pt}
\subsection*{Code availability} 
\url{https://bieqa.github.io/deeplearning.html}

\section{Introduction}

Graph convolutional neural networks (graph-CNNs) are deep learning techniques that apply to graph-structured data. Graph-structured data are in general complex, which imposes significant challenges on existing convolutional neural network algorithms. \red{Graphs are irregular and have variable number of unordered vertices with different topology at each vertex. This makes important algebraic operations such as convolutions and pooling challenging} to apply to the graph domain. Hence, existing research on graph-CNN has been focused on defining convolution and pooling operations. 

There are two types of approaches for defining convolution on a graph: one through the spatial domain and the other through the spectral domain \cite{bronstein.2017,zhang2020deep}. Existing \red{spatial} approaches, such as diffusion-convolutional neural networks (DCNNs) \cite{atwood2015diffusion}, PATCHY-SAN \cite{niepert2016learning,duvenaud2015conv}, gated graph sequential neural networks \cite{yujia2015gated}, DeepWalk \cite{perozzi2014deepwalk}, message-passing neural network (MPNN) \cite{gilmer2017neural}, develop convolution in different ways to process the vertices on a graph whose neighborhood has different sizes and connections. An alternative approach is to take into account of the geometry of a graph and to map individual patches of a graph to a representation that is more amenable to classical convolution, including 2D polar coordinate representation \cite{masci2015geodesic}, local windowed spectral representation \cite{boscaini2015learning}, anisotropic variants of heat kernel diffusion filters \cite{boscaini2016anisotropic,boscanini2016lscwacnn}, Gaussian Mixture-model kernels \cite{federico2016geometric}. 

On the other hand, several graph-CNN methods \red{called "spectral graph-CNN" defines convolution in the spectral domain}\cite{bruna2013spectral,Defferrard2016,henaff2015deep,kipf2016semi,liy2016syncspeccnn,ktena2017distance,shuman2016vertex}. The advantage of spectral graph-CNN methods lies in the analytic formulation of the convolution operation. Based on \red{the} spectral graph theory, Bruna et~al. \cite{bruna2013spectral} proposed convolution on graph-structured data in the spectral domain \red{via}  the graph Fourier transform. However, the eigendecomposition of the graph Laplacian for building the graph Fourier transform is computationally intensive when a graph is large. Moreover, spectral filters in \cite{bruna2013spectral} are non-localized in the spatial domain. Defferrard et~al. \cite{Defferrard2016} addressed these problems by proposing Chebyshev polynomials to parametrize spectral filters such that the resulting convolution is approximated by the polynomials of the graph Laplacian. Kipf and Welling \cite{kipf2016semi} adopted \red{the} first-order polynomial filter and stacked more spectral convolutional layers to replace higher-order polynomial \red{expansions.} In \cite{Defferrard2016,shuman2016vertex}, it is shown that the $k$-order Chebyshev polynomial approximation of graph Laplacian filters performs the $k$-ring filtering operation. 

In this study, we revisited the spectral graph-CNN based on the graph Laplacian \cite{Defferrard2016,shuman2016vertex} and developed the LB-CNN where spectral filters are designed via the Laplace-Beltrami (LB) operator on a graph. We call these filters as LB-based spectral filters. We investigated whether 
the proposed LB-CNN is superior to the graph-CNN \cite{Defferrard2016,shuman2016vertex} because the LB operator incorporates the intrinsic geometry of a graph \red{better than graph Laplacian} \cite{perrault2017improved}. We further explored the feasibility of Chebyshev, Laguerre, and Hermite polynomials to approximate LB-based spectral filters in the LB-CNN. We chose Laguerre and Hermite polynomials beyond Chebyshev polynomials since these polynomials have potentials to approximate the heat kernel convolution on a graph as shown in \cite{tan.2015,huang2020fast}. We employed the brain image data from Alzheimer's Disease Neuroimaging Initiative (ADNI) and demonstrated the use of the proposed LB-CNN. We compared the computational time and classification performance of the LB-CNN with the spectral graph-CNN \cite{Defferrard2016,shuman2016vertex} when Chebyshev, Laguerre, and Hermite polynomials were used. 
 
This study contributes to 
\begin{itemize}
\item providing the approximation of LB spectral filters using Chebyshev, Laguerre, Hermite polynomials and their implementation in the LB-CNN;
\item updating the LB operator for pooling in the LB-CNN;
\item demonstrating the feasibility \red{of using} the LB operator and different polynomials for graph-CNNs.
\end{itemize}

\section{Methods}
\label{sec:methods}
Similar to classical CNN and spectral graph-CNN \cite{Defferrard2016}, the LB-CNN has three major components, including convolution, rectified linear unit (ReLU), and pooling. In the following, we will first describe LB spectral filters in the convolutional layer and how to define a pooling operation via coarsening a graph and updating the LB operator.

\subsection{Laplace-Beltrami spectral filters}
\subsubsection{Polynomial approximation of LB spectral filters}
\red{Consider the Laplace-Beltrami (LB) operator $\Delta$ on surface $\mathcal{M}$.} Let $\psi_j$ be the $j^{th}$ eigenfunction of the LB-operator with eigenvalue $\lambda_j$
\begin{equation}\label{eq:eigs}
{\Delta}\psi_j=\lambda_j\psi_j \ ,
\end{equation}
where $0=\lambda_0 \leq \lambda_1 \leq \lambda_2 \leq \cdots$. A signal $f(x)$ on the surface $\mathcal{M}$ can be represented as a linear combination of the LB eigenfunctions
\begin{equation}\label{eq:x}
f(x)=\sum_{j=0}^{\infty}c_j\psi_j(x) \  ,
\end{equation}
where $c_j$ is the $j^{th}$ coefficient associated with the eigenfunction $\psi_j(x)$. 

\begin{figure}[t]
\centering
\includegraphics[width=1\linewidth]{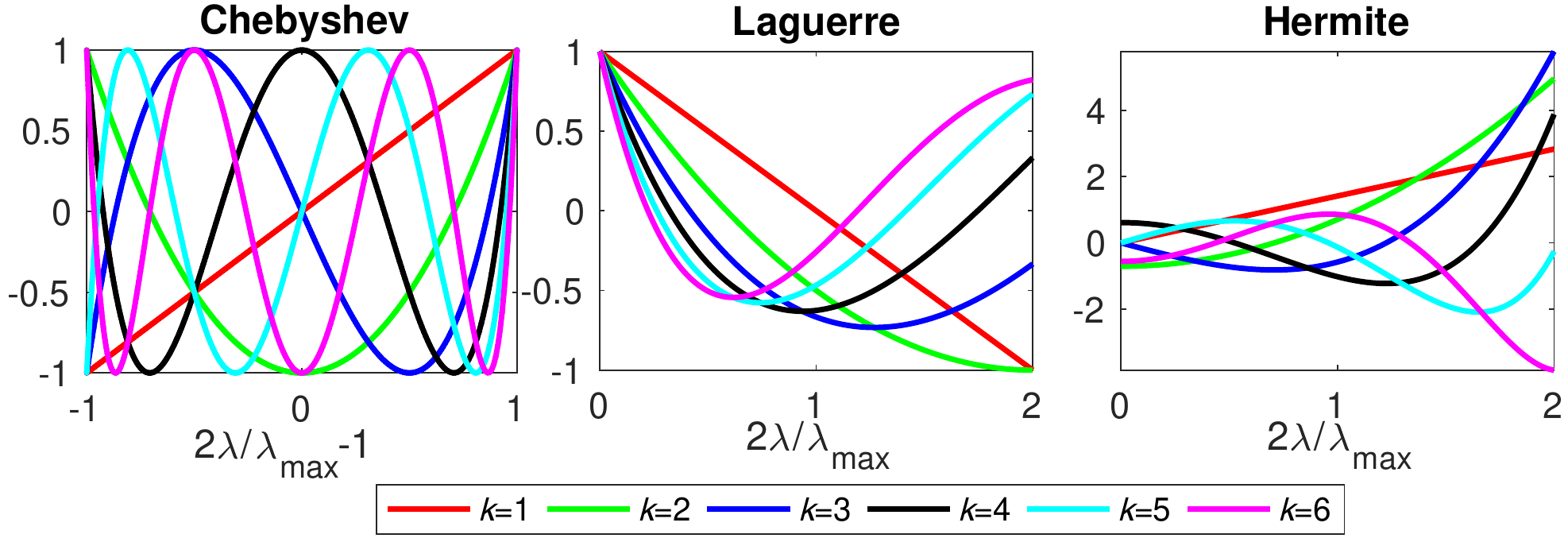}
\caption{Chebyshev, Laguerre and Hermite polynomials of order from 1 to 6. The eigenvalues were scaled and shifted to $[-1,1]$  for Chebyshev polynomials and to $[0,2]$ for Laguerre and Hermite polynomials, where $\lambda_{\max}$ is the maximum eigenvalue of the LB-operator.}
\label{fig:spectral}
\end{figure}

We now consider an LB spectral filter $g$ on $\mathcal{M}$ with spectrum $g(\lambda)$ as
\begin{equation}\label{eq:g}
g(x,y)=\sum_{j=0}^{\infty}g(\lambda_j)\psi_j(x)\psi_j(y) .
\end{equation}
Based on Eq. (\ref{eq:x}), the convolution of a signal $f$ with the filter $g$ can be written as  
\begin{equation}\label{eq:gx}
 h(x) = g \ast f(x)
=\sum_{j=0}^{\infty}g(\lambda_j) c_j\psi_j(x).
\end{equation}
As suggested in \cite{Defferrard2016,wee2019cortical,coifman2006diffusion,Hammond2011129,kim2012wavelet,tan.2015}, the filter spectrum $g(\lambda)$ in  Eq. (\ref{eq:gx}) can be approximated as the expansion of Chebyshev polynomials, $T_k$, 
$k=0, 1, 2, \dots, K-1$, such that
\begin{equation}
\label{eq:glambda}
g(\lambda)=\sum_{k=0}^{K-1}\theta_k T_k(\lambda) \ .
\end{equation}
$\theta_k$ is the $k ^{th}$ expansion coefficient associated with the $k ^{th}$ Chebyshev polynomial.  $T_k$ is the Chebyshev polynomial of the form
$T_k(\lambda)=\cos(k\cos^{-1}\lambda)$.
The left panel on Fig. \ref{fig:spectral} shows the shape of the $k^{th}$ Chebyshev polynomial up to order 6.  
We can rewrite the convolution in Eq. (\ref{eq:gx}) as 
\begin{equation}\label{eq:gx3}
h(x) = g \ast f(x)=\sum_{k=0}^{K-1}\theta_k T_k(\Delta) f(x).
\end{equation}

Likewise, $g(\lambda)$ in  Eq. (\ref{eq:gx}) can also be approximated using other polynomials, such as Laguerre or Hermite polynomials \cite{olver2010nist}. $T_k$ in Eq. (\ref{eq:gx3}) can be replaced by Laguerre, $L_k$, or Hermite, $H_k$, polynomials, where 
\begin{align} \label{eq:TLH}
L_k(\lambda)&=\sum_{l=0}^k \binom{k}{l}\frac{(-\lambda)^l}{l!} ,\\
H_k(\lambda)&=k!\sum_{l=0}^{\lfloor k/2 \rfloor}\frac{(-1)^l(2\lambda)^{k-2l}}{l!(k-2l)!} \ , \nonumber 
\end{align}
In this paper, we adopt the following normalized definition of Hermite polynomials:
\begin{equation}\label{eq:normH}
\bar{H}_k(\lambda)=\frac{1}{\sqrt{2^kk!}}H_k(\lambda) \,
\end{equation}
where the inner product of $\bar{H}_k$ with itself is independent of $k$. The last two panels of Fig. \ref{fig:spectral} show the shapes of Laguerre and Hermite polynomials up to order 6, respectively. 


\subsubsection{Numerical implementation of LB spectral filters via polynomial approximations}

We now discretize the surface $\mathcal{M}$ as a triangulated mesh, $\mathcal{G}=\{V, E\}$, with  a set of triangles and vertices $v_i$. For the implementation of the LB spectral filters, we adopt the discretization scheme of the LB operator in \cite{tan.2015}. The $ij^{th}$ element of the LB-operator on $\mathcal{G}$ can be computed as 
\begin{equation}
\label{eq:Delta}
\Delta_{ij}= C_{ij} /A_i,
\end{equation}
where $A_i$ is 
estimated by the Voronoi area of nonobtuse triangles \cite{meyer2003discrete} and the Heron's area of obtuse triangles containing $v_i$  \cite{tan.2015,meyer2003discrete}.
The off-diagonal entries are defined as $C_{ij}=-(\cot\theta_{ij}+\cot\phi_{ij})/2$ if $v_i$ and $v_j$ form an edge, otherwise $C_{ij}=0$. The diagonal entries $C_{ii}$  are computed as $C_{ii}=-\sum_{j} C_{ij}$. Other cotan discretizations of the LB operator are discussed in \cite{chung.2004.ISBI,qiu.2006, chung.2015.MIA}. When the number of vertices on $\mathcal{M}$ is large, the computation of the LB eigenfunctions can be costly \cite{huang2020fast}.

For the sake of simplicity, we denote the $k^{th}$ order polynomial as $P_k$, where $P_k$ can represent Chebyshev, Laguerre, or Hermite polynomial. We take the advantage of the recurrence relation of these polynomials (Table~\ref{tab:orth_recurr}) and compute LB spectral filters recursively as follows.
\begin{itemize}
  \item[1.] compute $\Delta$ based on Eq. (\ref{eq:Delta}) for the triangulated mesh $\mathcal{G}$;
  \item[2.] compute the maximum eigenvalue $\lambda_{\max}$ of $\Delta$. For the standardization across surface meshes, we normalize $\Delta$ as $\widetilde\Delta=\frac{2\Delta}{\lambda_{\max}}-I$ such that the eigenvalues are mapped from $[0,\lambda_{\max}]$ to $[-1,1]$ for Chebyshev polynomials \cite{Defferrard2016,huang2020fast}. $I$ is an identity matrix. For Laguerre and Hermite polynomials, we normalize $\Delta$ as $\widetilde{\Delta}=\frac{2\Delta}{\lambda_{\max}}$, which maps the eigenvalues from $[0,\lambda_{\max}]$ to $[0,2]$;

 \item[3.] for a signal $f(x)$, compute  $P_k(\widetilde\Delta)f(x)$ recursively by
\begin{equation}\label{eq:recurrence_Delta}
P_{k+1}({\widetilde\Delta}) f= A_k{\widetilde\Delta}\  P_{k}({\widetilde\Delta}) f+B_kP_{k}({\widetilde\Delta}) f+ C_k P_{k-1}({\widetilde\Delta}) f,
\end{equation}
with the initial conditions $P_{-1}({\widetilde\Delta}) f(x)=0$ and $P_0({\widetilde\Delta})  f(x) = f(x)$. The recurrence relations of different polynomials are given in Table~\ref{tab:orth_recurr}.
\end{itemize}
Step 3 is repeated from $k=0$ till $k=K-2$. 

\begin{table}[t]
\caption{The recurrence relation of Chebyshev, Laguerre and Hermite polynomials in spectral filtering.}\label{tab:orth_recurr}
\footnotesize{\setstretch{2.2}
\begin{tabular}{| l | l | }
\hline
Method & Recurrence relations\\
\hline
Chebyshev$^\dagger$ & $T_{k+1}(\widetilde{\Delta}) f= (2-\delta_{k0})\widetilde{\Delta} \ T_{k}(\widetilde{\Delta}) f- T_{k-1}(\widetilde{\Delta})  f$ \\
\hline
Laguerre & $L_{k+1}(\widetilde{\Delta}) f= \frac{-\widetilde{\Delta} \ L_{k}(\widetilde{\Delta})  f+(2k+1)L_{k}(\widetilde{\Delta})  f- kL_{k-1}(\widetilde{\Delta}) f}{k+1} $ \\
\hline
Hermite & $\bar{H}_{k+1}(\widetilde{\Delta}) f= \sqrt{\frac{2}{k+1}}\widetilde{\Delta} \ \bar{H}_{k}(\widetilde{\Delta})  f- \sqrt{\frac{k}{k+1}}\bar{H}_{k-1}(\widetilde{\Delta})  f$  \\
\hline
\end{tabular}}
\blue{\small{$^\dagger$ $\delta_{k0}$ is Kronecker delta.}}
\end{table}
 
\subsubsection{Localization of spectral filters based on polynomial approximations}
Analogue to the spatial localization property of Chebyshev polynomial approximation of graph Laplacian spectral filters \cite{Defferrard2016}, we can show that Chebyshev, Laguerre, or Hermite polynomial approximation of LB spectral filters also has this localization property. We consider the discretization of $\Delta$ given in Eq. (\ref{eq:Delta}).  Consider  two vertices  $v_i$ and $v_j$ on $\mathcal{G}$. 
We can define the shortest  distance  between $v_i$ and $v_j$, denoted by $d_\mathcal{G}(i,j)$,  as the minimum number of edges on the path connecting $v_i$ and $v_j$. 
Hence, 
\begin{equation}
\label{eq:lemma5.2}
(\Delta^K)_{i,j}=0 \ \ \textmd{ if }\ \ d_\mathcal{G}(i,j)>K,
\end{equation}
where $\Delta^K$ denotes \red{the $K$-th} power of the LB operator $\Delta$ \cite{tan.2015}.  In other words, the coverage of $(\Delta^K)_{i,j}$ is localized in the ball with radius $k$ from the central vertex. 

$P_k(\Delta)$ can be represented in terms of $\Delta^0, \Delta, \cdots, \Delta^k$ and is $k$-localized $(P_k(\Delta))_{i,j}=0$ if $d_\mathcal{G}(i,j)>k$ according to Eq. (\ref{eq:lemma5.2}).
The spectral filter $g$ composed of $P_0(\Delta)$, $P_1(\Delta)$, ..., $P_{K-1}(\Delta)$
is a spatially localized filter with localization property given by
\begin{equation}\label{eq:local}
\left(g(\Delta)\right)_{i,j}=0 \ \ \textmd{ if }\ \ d_\mathcal{G}(i,j)>K-1.
\end{equation}

In practice, we can also show the spatial localization of filter $g$ composed of $P_0(\Delta)$, $P_1(\Delta)$, ..., $P_{K-1}(\Delta)$ by applying $g$ to an impulse signal $f_j$ with $1$ at vertex $v_j$  and $0$ at the others. Then, the filter output is given by $g(x,v_j)=\sum_{k=0}^{K-1}\theta_k P_k(\Delta) f_j(x)$.
When $x=v_i$ satisfying $d_\mathcal{G}(i,j)>K-1$, since $(P_k(\Delta))_{i,j}=0$, we have
\begin{equation}\label{eq:local2}
g(v_i,v_j)=\sum_{k=0}^{K-1}\theta_k \left(P_k(\Delta) f_j(x)\right)_i=\sum_{k=0}^{K-1}\theta_k \left(P_k(\Delta)\right)_{ij}=0.
\end{equation}


\subsection{Rectified Linear Unit}
Similar to classic CNN, a \emph{rectified linear unit} (ReLU) in the LB-CNN can be represented by many non-linear activation functions. The activation function is a map from $\RR$ to $\RR$, which does not involve any geometrical property of a triangulated mesh. In our proposed LB-CNN on a mesh, we adopt the well-known ReLU:
$$
\sigma(z)=\max\{0, z\},\quad z\in \RR.
$$ 

\subsection{Mesh coarsening and pooling}
For the LB-CNN, the pooling layer involves mesh coarsening, pooling of signals, and an update of the LB operator. First, we adopt the Graclus multilevel clustering algorithm \cite{dhillon2007weighted} to coarsen a graph based on the graph Laplacian. This algorithm is built on the METIS \cite{karypis1998fast} to cluster similar vertices together from a given graph by a greedy algorithm. At each coarsening level, two neighboring vertices with maximum local normalized cut are matched until all vertices are explored \cite{shi_pami_2000}. In our case,  the discrete LB-operator $\Delta$ in Eq. (\ref{eq:Delta}) is used.  The local normalized cut on a mesh is computed by $-\Delta_{ij}(1/\Delta_{ii}+1/\Delta_{jj})$.
The coarsening process is repeated until the coarsest level is achieved. After coarsening, a balanced binary tree is generated where each node has either one (i.e. singleton) or two child nodes. Fake nodes are added to pair with those singleton. The weights of the edges involving fake nodes are set as 0.  Then, the pooling on this binary tree can be efficiently implemented as a simple 1-dimensional pooling of size 2.

We now discuss the update of the LB operator for a coarsen mesh. When two matched nodes are merged together at a coarser  level, the weights of the edges involving the two nodes are merged by summing them together. By doing so, each coarsened mesh has its updated $\Delta$.

\subsection{LB-CNN Architecture}
 \begin{figure}[t]
\centering
\includegraphics[width=1\linewidth]{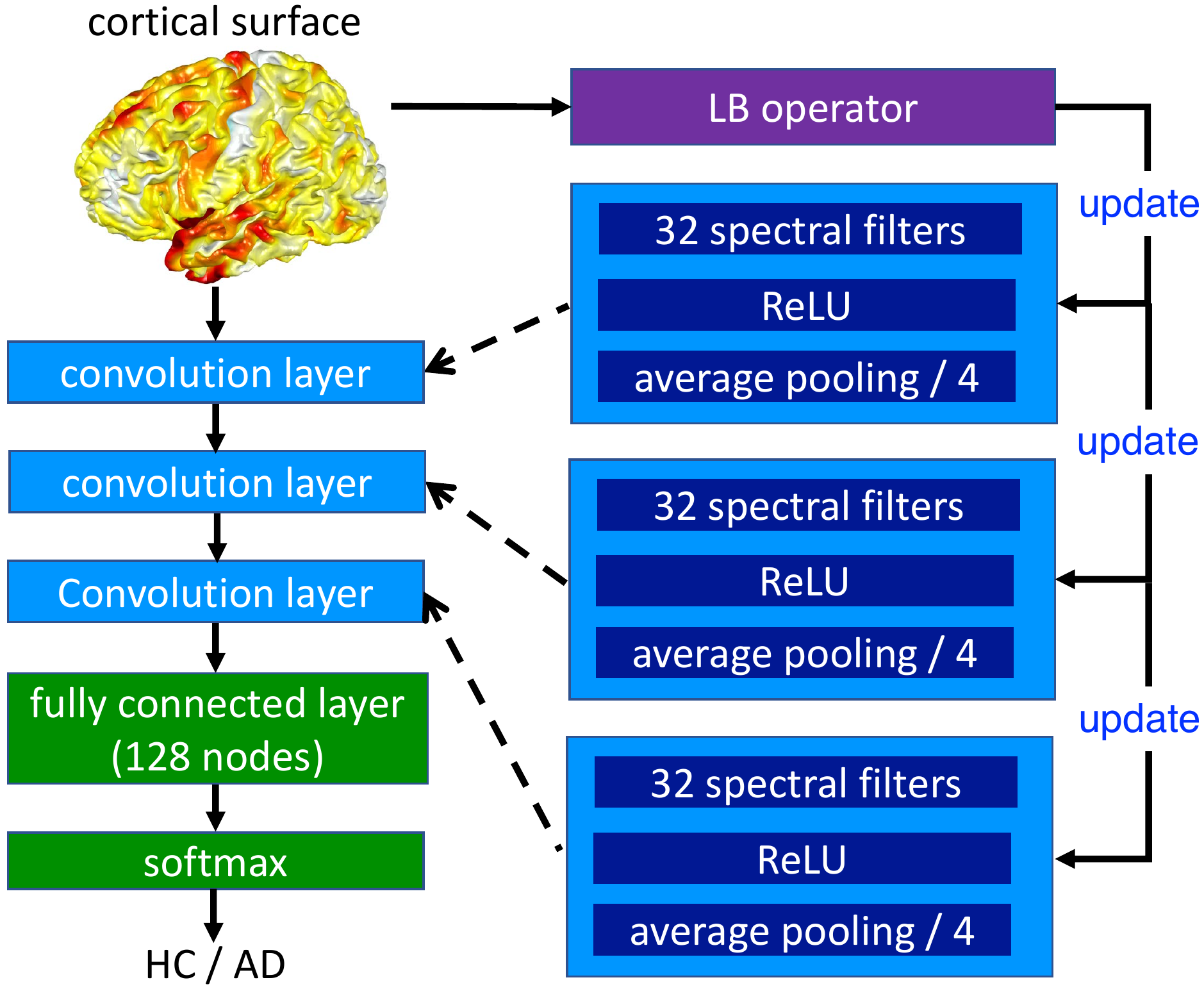}
\caption{An example of LB-CNN architectures for the classification of health controls (HC) and \red{Alzheimer's Disease (AD) patients}. The cortical surface is represented as a triangulated mesh with 655,360 triangles and 327,684 vertices.
}
\label{fig:architecture}
\end{figure}

We are now well equipped with all the components for a LB-CNN network. The LB-CNN network is composed of total 
$L+1$ connected stages. The first $L$ stages are the stages for feature extraction. Each stage contains three sequentially concatenated layers: (1) a convolutional layer with multiple LB spectral filters; (2) a ReLU layer;  (3)  a pooling layer with stride 2 or higher that uses average pooling. 
In the last stage, a fully connected layer followed by a softmax function is employed to make a decision, and the output layer contains classification labels.

Fig. \ref{fig:architecture} illustrates one of LB-CNN architectures that are analogous to classical CNN for image data defined on equi-spaced grids.  In this example, the $i$-th convolution layer is composed of $2^{i+2}$ LB spectral filters that can be approximated using Chebyshev, Laguerre, and Hermite polynomials, an ReLU,  and an average pooling  with pooling size $2^{\max\{5-i,1\}}$ and  stride being the same as the pooling size. In the fully connected layer, there are $128$ hidden nodes, and an $l_2$-norm regularization with weight of $5\times 10^{-4}$ is applied to prevent overfitting.

All the networks can be trained by \red{the} back propagation algorithm with $30$ epochs, mini-batch size of $32$, initial learning rate of $10^{-3}$, learning rate decay of $0.05$ for every $20$ epochs, momentum of $0.9$ and no dropout.

\subsection{MRI data acquisition and preprocessing}
We utilized the structural T1-weighted MRI from the ADNI-2 cohort (\protect\url{adni.loni.ucla.edu}). 
The aim of this study was to illustrate  the use of the LB-CNN and spectral graph-CNN via the HC/AD classification since it has been well studied using T1-weighted image data (e.g., 
\cite{cuingnet2011automatic,liu2013locally,hosseini2016alzheimer,korolev2017residual,liu2018landmark,islam2018brain,basaia2019automated,wee2019cortical}). 
Hence, this study involved 643 subjects with HC or AD scans (392 subjects had HC scans; 253 subjects had AD scans). There were 8 subjects who fell into both groups due to the conversion from HC to AD.  There were total 1122 scans for HC and 587 for AD.

The MRI data of the ADNI-2 cohort were acquired from participants aging from 55 to 90 years using either 1.5 or 3T scanners. 
The T1-weighted images were segmented using FreeSurfer (version 5.3.0) \cite{fischl2002whole}. The white and pial cortical surfaces were generated at the boundary between white and gray matter and the boundary of gray matter and CSF, respectively. Cortical thickness was computed as the distance between the white and pial cortical surfaces. It represents the depth of the cortical ribbon. We represented cortical thickness on the mean surface, the average between the white and pial cortical surfaces.  We employed large deformation diffeomorphic metric mapping (LDDMM) \cite{zhong2010quantitative,du2011whole}  to align individual cortical surfaces to the atlas and transferred the cortical thickness of each subject to the atlas. The cortical atlas surface was represented as a triangulated mesh with 655,360 triangles and 327,684 vertices. At each surface vertex, a  spline regression implemented by piecewise step functions \cite{james2013introduction} was performed to regress out the effects of age and gender. The residuals from the regression were used in the below spectral graph-CNN and LB-CNN.

\section{Results}
\label{sec:results}

\subsection{Spatial localization of the LB spetral filters via polynomial approximations}

\begin{figure}[t]
\centering
\includegraphics[width=1\linewidth]{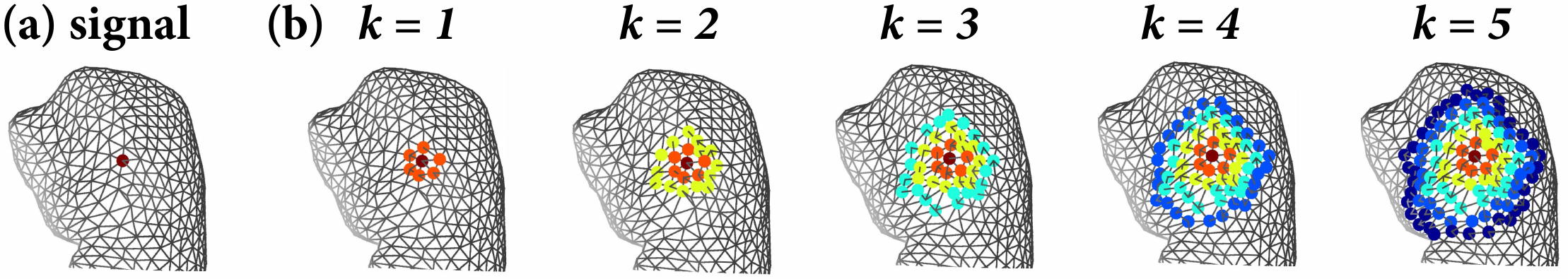}
\caption{The first panel shows the signal with 1 at one vertex and 0 at the other vertices of the hippocampus.  The rest of panels shows the spatial localization of spectral filters using Chebyshev $T_k$, Laguerre  $L_k$ and Hermite  $\bar{H}_k$ polynomials for $k=1,2,...,5$. }
\label{fig:local}
\end{figure}

\begin{figure}[t]
\centering
\includegraphics[width=1\linewidth]{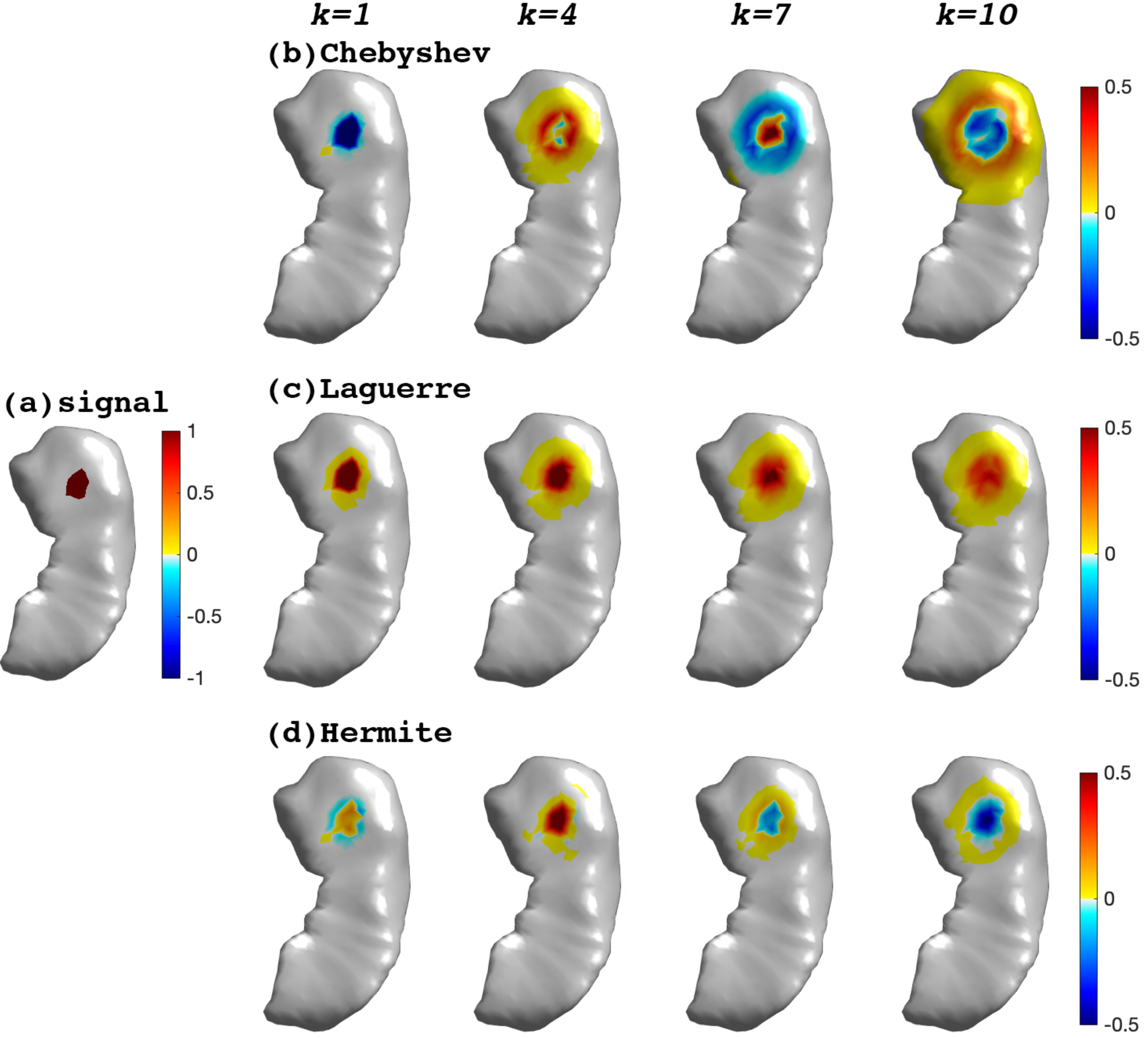}
\caption{The first panel shows the input signal, where 1 is on a red region and 0 on the rest of the hippocampus. The rest panels show the signals after filtering via LB spectral filters, $g=T_k$, $L_k$ and $\bar{H}_k$ for $k=1,4,7,10$, where $T_k$, $L_k$ and $\bar{H}_k$ are  the Chebyshev, Laguerre, and Hermite polynomials, respectively.
}
\label{fig:spatial}
\end{figure}

Fig.~\ref{fig:local}  shows the localization property of spectral filters using the Chebyshev, Laguerre and Hermite polynomials. The input signal is 1 at only one vertex and 0 at all other vertices of the hippocampus. The  $P_k(\Delta)$ is strictly localized in a ball of radius $k$, i.e., $k$ rings from the central vertex. Fig.~\ref{fig:spatial}

Consider a signal having 1 on a small patch (see the first panel in Fig.~\ref{fig:spatial}) and 0 on the rest of a hippocampus surface mesh with 1184 vertices and 2364 triangles. Fig.~\ref{fig:spatial}  shows the convolutions of this signal with spectral filters $g=T_k$, $L_k$ or $\bar{H}_k$ for $k=1,4,7,10$.
The spectral filters designed by different polynomials show different impacts on the signal in the spatial domain.

\subsection{Comparison of spectral graph-CNN and LB-CNN}
\begin{table*}[t]
\caption{Classification performance of the spectral graph-CNN and LB-CNN with Chebyshev, Laguerre, and Hermite polynomial approximations.} 
\label{tab:ACC}
\center{\begin{tabular}{ | l | l | c | c | c c c  c   |}
\hline
Spectral CNN & Polynomial &  Layer &  $K$    & ACC (\%) & SEN (\%) & SPE (\%)  & GMean (\%) \\ 
\hline
\hline
\multirow{3}{*}{Graph}  & {Chebyshev} & 4 & 6 
& $89.8\pm0.4$	& $90.1\pm0.9$ &	 $89.6\pm0.6$ & $89.9\pm0.4$ 	\\
&{Laguerre} & 5 & 7 & $90.0\pm0.5	$	& $91.7\pm1.1$ &	 $89.2\pm0.6$ & $90.4\pm0.6$	\\
&{Hermite}  & 3 & 7 &$87.1\pm0.5$ 	& $86.6\pm1.6$ &	 $87.4\pm0.9$ & $87.0\pm0.7$	 \\
\hline
\multirow{3}{*}{LB}  & {Chebyshev} & 5& 7 
& $90.9\pm0.6$	& $91.3\pm0.1$ &	 $90.7\pm0.5$ & $91.0\pm0.7$  \\
&{Laguerre} & 5 & 7 & $91.0\pm0.4$	& $91.2\pm0.9$ &	 $90.9\pm0.8$ & $91.1\pm0.4$	 \\
&{Hermite}  & 4 & 7 &$88.2\pm0.6$ 	& $87.5\pm0.6$ &	 $88.4\pm1.1$ & $88.0\pm0.4$	 \\
\hline
\end{tabular}}\\
\small{ACC: accuracy; SEN: sensitivity; SPE: specificity; GMean: geometric mean.}
\end{table*}

We aimed to compare the computational cost and classification accuracy of the spectral graph-CNN \cite{Defferrard2016,wee2019cortical} and LB-CNN  on the cortical thickness of the HC and the AD patients while Chebyshev, Laguerre and Hermite polynomials were used to approximate spectral filters. 

In our experiments, the architecture of the spectral graph-CNN and LB-CNN was the same as shown in Fig. \ref{fig:architecture} except the number of layers.  Ten-fold cross-validation was applied to the dataset (HC: $n = 1122$; AD: $n = 587$). One fold of real data was left out for testing. The remaining nine folds were further separated into training ($75\%$) and validation ($25\%$) sets randomly. 
To prevent potential data leakage in the ten-fold cross-validation, we constructed non-overlap training, validation, and testing sets  with respect to subjects instead of MRI scans. This ensured that the scans from the same subjects were in the same set.
The above data splitting was done  for the HC and AD groups separately so that the ratio of the number of subjects in the two groups was similar in all sets.

\begin{figure}[t]
\centering
\includegraphics[width=1\linewidth,clip=true]{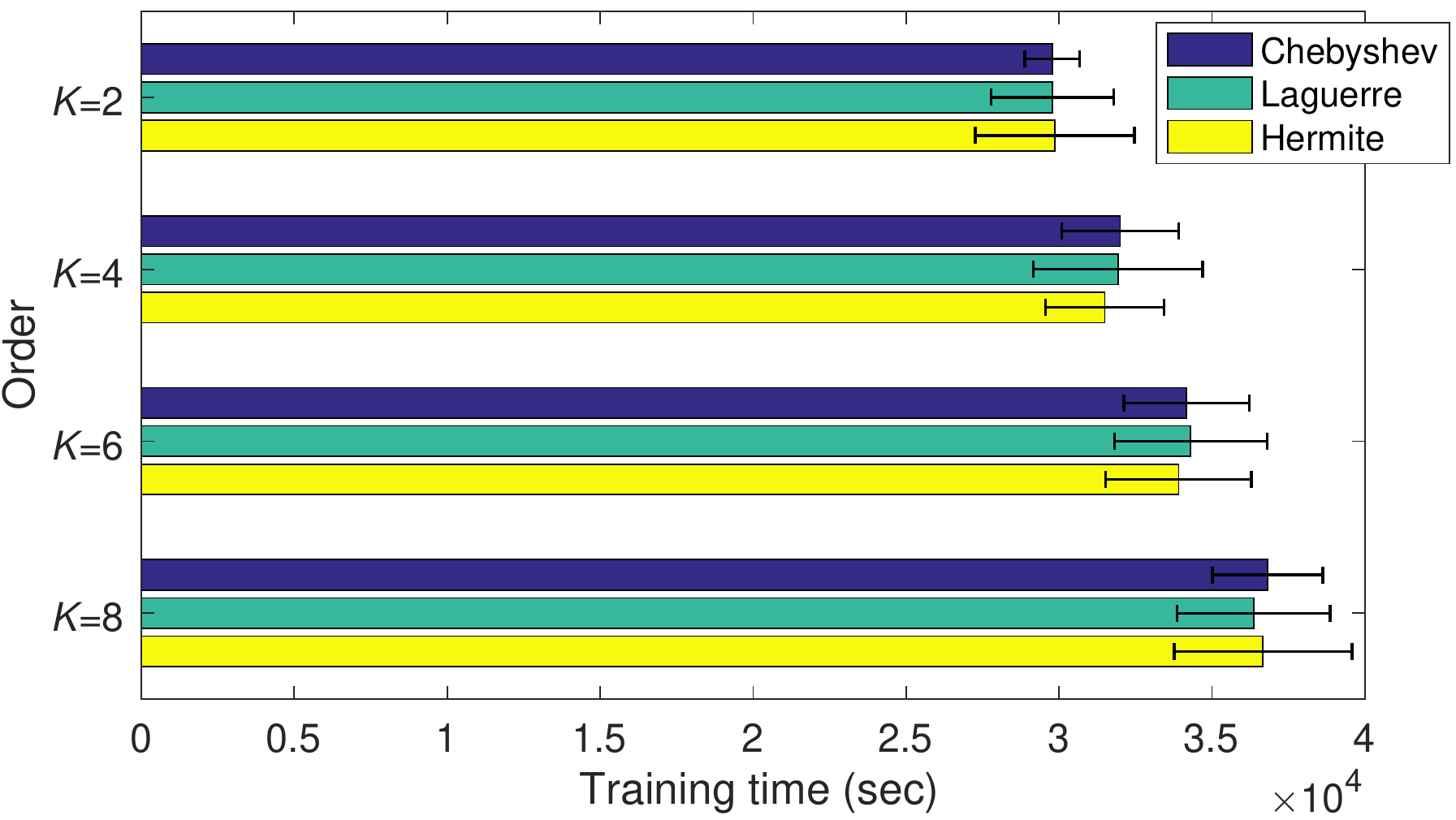}
\caption{
Computational cost of the LB-CNN using Chebyshev (blue), Laguerre (green) and Hermite (yellow) approximations of different order $K$. 
The ten-fold cross-validation was repeated for 10 times. The average and standard deviation of the training time are shown.}
\label{fig:Time}
\end{figure}

\subsubsection{Computational Cost}
The computation cost of the LB-CNN was similar to that of the spectral graph-CNN \red{since they only differ on the 
edge weights between vertices.} For instance, when the network with 3 convolutional layers and 6-order Hermite approximation was used,  the spectral graph-CNN and LB-CNN respectively had training time $33625 \pm 1310$ sec and $33908 \pm 2385$ sec over the ten-fold cross validation with no significant difference (two-sample $t$-test $p=0.75$).

Next, we compared the computational cost of the LB-CNN with 3 convolutional layers using different polynomial approximations. Fig.~\ref{fig:Time} shows the training time of the LB-CNNs using the Chebyshev, Laguerre and Hermite approximation of order $K=2$, $4$, $6$ and $8$. Regardless of which polynomial was used, the training time increased as $K$ increased since more trainable parameters were needed to characterize the spectral filters. Given $K$, the three polynomial approximation methods had similar computation cost ($p>0.56$). 

\subsubsection{Classification Performance}
To compare classification performance of the spectral graph-CNN and LB-CNN on HC and AD, the number of convolutional layers and polynomial approximation order were tuned for each CNN independently to achieve the best classification accuracy and geometric mean (GMean) on the validation set. The spectral graph-CNNs with Chebyshev, Laguerre and Hermite approximations respectively required 4 convolutional layers with polynomial order of $K=6$, 5 layers with $K=7$, and 3 layers with $K=7$.
The LB-CNNs with Chebyshev and Laguerre approximations needed 5 layers with $K=7$
while the LB-CNN with Hermite approximation required 4 layers with $K=7$.
Table~\ref{tab:ACC} lists the accuracy, sensitivity, specificity and GMean of all these CNNs in classifying AD and HC.

Two-sample $\it{t}$-test found no significant difference in classification accuracy between the spectral graph-CNN and LB-CNN. For instance, when Chebyshev polynomials were used to approximate the spectral filters, the spectral graph-CNN classification accuracy was $89.9\%$ while the LB-CNN classification accuray was $90.9\%$ ($p=5.4\times10^{-5}$). Likewise, there were no group differences in classification accuracy between the spectral graph-CNN and LB-CNN when Laguerre and Hermite polynomial approximations were used (Laguerre: $p=3.5\times10^{-4}$; Hermite:$p=9.5\times10^{-4}$). Hence, the LB- CNN classification performance \red{can be viewed as} equivalent to the spectral graph-CNN.

As for the comparisons among the three different polynomials,  the classification performance of the Laguerre approximation was comparable to the Chebyshev approximation (graph-CNN: $p=0.27$; LB-CNN: $p=0.81$). \red{However,} the classification performance of both Chebyshev and Laguerre polynomial approximations was greater than that of the Hermite polynomial approximation (all $p<5.4\times10^{-10}$). In \cite{huang2020fast}, Hermite polynomial approximation  shows slower convergence to heat kernel, compared to Chebyshev and Laguerre polynomial approximations.

%

\section{Discussions}
In this study, we revisited the spectral graph-CNN \cite{Defferrard2016,shuman2016vertex} and developed the LB-CNN by replacing the graph Laplacian by the LB operator.  We also employed Chebyshev, Laguerre, and Hermite polynomials to approximate the LB spectral filters in the LB-CNN and spectral graph-CNN. Based on cortical thickness of the ADNI dataset, we showed that the LB-CNN didn't improve classification accuracy compared to the spectral graph-CNN \cite{Defferrard2016,shuman2016vertex}. The three polynomials had the similar computational cost and showed comparable classification accuracy in the LB-CNN or spectral graph-CNN \cite{Defferrard2016,shuman2016vertex}. Our findings suggest that even though the shapes of the three polynomials are different, deep learning architecture allows to learn spectral filters such that the classification performance is not dependent on the type of the polynomials or the operators (graph Laplacian and LB operator).

\bibliographystyle{spbasic}      
\bibliography{GCNN_arxiv}

\end{document}